\documentclass{article}
\usepackage{iclr2020_conference,times}

\usepackage{amsmath,amsfonts,bm}

\def\eqref#1{equation~\ref{#1}}
\def\1{\bm{1}}

\DeclareMathAlphabet{\mathsfit}{\encodingdefault}{\sfdefault}{m}{sl}
\SetMathAlphabet{\mathsfit}{bold}{\encodingdefault}{\sfdefault}{bx}{n}

\usepackage{hyperref}
\usepackage{url}
\usepackage{booktabs}
\usepackage[utf8]{inputenc}
\usepackage{amsfonts}
\usepackage{nicefrac}
\usepackage{mathtools}
\usepackage{microtype}
\usepackage{color}
\usepackage{graphicx}
\usepackage{makecell}
\usepackage{caption}
\usepackage{subfig}
\usepackage{comment}
\usepackage{float}
\usepackage[T1]{fontenc}

\title{Understanding the robustness of deep neural network classifiers for breast cancer screening}

\author{Witold Oleszkiewicz$^{5,*}$, Taro Makino$^{1,*}$, Stanis\l{}aw Jastrz\k{e}bski$^{ 1,3,*}$, Tomasz Trzci\'nski$^5$,\\
\textbf{Linda Moy$^{1,2,6}$, Kyunghyun Cho$^{3,4,7}$, Laura Heacock$^{1,6}$, Krzysztof J. Geras$^{1,2,3}$}\\
\\
$^1$Department of Radiology, NYU Grossman School of Medicine\\
$^2$Center for Advanced Imaging Innovation and Research, NYU Langone Health\\
$^3$Center for Data Science, New York University\\
$^4$Department of Computer Science, Courant Institute, New York University\\
$^5$Faculty of Electronics and Information Technology, Warsaw University of Technology\\
$^6$Perlmutter Cancer Center, NYU Langone Health\\
$^7$CIFAR Associate Fellow\\
*Equal contribution
}

\newcommand{\nan}{\mathcal{M}_{dmv}}
\newcommand{\artie}{\mathcal{M}_{gmic}}

\iclrfinalcopy

\begin{document}

\maketitle

\begin{abstract}
    Deep neural networks (DNNs) show promise in breast cancer screening, but their robustness to input perturbations must be better understood before they can be clinically implemented. There exists extensive literature on this subject in the context of natural images that can potentially be built upon. However, it cannot be assumed that conclusions about robustness will transfer from natural images to mammogram images, due to significant differences between the two image modalities. In order to determine whether conclusions will transfer, we measure the sensitivity of a radiologist-level screening mammogram image classifier to four commonly studied input perturbations that natural image classifiers are sensitive to. We find that mammogram image classifiers are also sensitive to these perturbations, which suggests that we can build on the existing literature. We also perform a detailed analysis on the effects of low-pass filtering, and find that it degrades the visibility of clinically meaningful features called microcalcifications. Since low-pass filtering removes semantically meaningful information that is predictive of breast cancer, we argue that it is undesirable for mammogram image classifiers to be invariant to it. This is in contrast to natural images, where we do not want DNNs to be sensitive to low-pass filtering due to its tendency to remove information that is human-incomprehensible.
\end{abstract}

\section{Introduction}

The clinical implementation of AI can improve healthcare accessibility both economically and geographically by reducing the workload of clinicians. In economically disadvantaged areas where clinicians are scarce, this means that patients can get care when they otherwise would not be able to. Benefits can also be expected in areas where clinicians are available, since they are subject to fatigue, which can make them more prone to error. Following their success in natural image object recognition~\citep{Krizhevsky2012AlexNet}, deep neural networks (DNNs) show great promise in breast cancer screening~\citep{Wu2019Mammography, Shen2019GMIC, Shen2020GMIC, Mckinney2020BreastCancer}. However, strong performance in specific evaluation settings is insufficient to consider DNNs ready for clinical deployment. This is because DNNs are vulnerable to distribution shifts that are seemingly innocuous from the perspective of human vision~\citep{Jo2017SurfaceStats, Geirhos2018TextureBias, Hendrycks2019CommonCorruptions, Yin2019Fourier}. This is among the most significant weaknesses that hinder the implementation of DNNs in safety-critical applications, and while being well-documented in the context of natural images, it has not yet been studied in the context of mammogram images. In order to understand the robustness of mammogram image classifiers, it would be greatly beneficial to be able to build on the existing literature. However, it is unclear whether conclusions about robustness will transfer from natural images to mammogram images, because in contrast to natural images, salient features in mammogram images can be tiny relative to the overall spatial resolution of the image.

This motivates the main contribution of our paper, which is the first study of DNN robustness in the context of screening mammography. More specifically, we analyze the robustness of two recently released radiologist-level screening mammogram image classifiers~\citep{Wu2019Mammography, Shen2019GMIC, Shen2020GMIC} with respect to Fourier high- and low-pass filtering, additive Gaussian noise, and patch shuffling; please see Figure~\ref{fig:image_perturbations} for examples. Our results show that mammogram image classifiers are sensitive to the four perturbations under consideration, meaning that further work on DNN robustness in screening mammography can build on an extensive existing body of work.

We also take a closer look at how mammogram image classifiers react to low-pass filtering, since this is particularly well-studied in the context of natural images. Our objective is to understand, in clinically interpretable terms, what information in mammogram images corresponds to high frequencies. We identify that low-pass filtering degrades the visibility of microcalcifications, which are important features that radiologists look for during breast cancer screening~\citep{Rominger2015}. The implication is that since low-pass filtering removes these clinically meaningful features, as long as mammogram image classifiers also utilize this information, their sensitivity to low-pass filtering is to be expected. In fact, since being invariant to low-pass filtering would hurt the ability of mammogram image classifiers to detect microcalcifications, we argue that some sensitivity is desirable in this context. This view is opposite to the consensus for natural images - since low-pass filtering typically removes information from natural images that is human-incomprehensible~\citep{Jo2017SurfaceStats,Ilyas2019NotBugs}, it is considered undesirable for natural image classifiers to be sensitive to it.

\section{Related work}
\label{sec:related work}

While our work is the first empirical study of DNN robustness in the context of screening mammography, the topic has been extensively studied for natural images from a wide range of perspectives. These include human-imperceptible adversarial attacks~\citep{Szegedy2014Intriguing}, Fourier filtering~\citep{Jo2017SurfaceStats}, and naturally occurring corruptions such as noise and blurring~\citep{Hendrycks2019CommonCorruptions}. While such distribution shifts break DNNs, humans are remarkably robust to them. A potential explanation for this is that DNNs discriminate using a different set of features than humans. Several authors have reached this conclusion while coming from different directions. For instance, \citet{Jo2017SurfaceStats} apply Fourier filtering to show that DNNs trained on natural images are sensitive to human imperceptible low-pass filtering. By demonstrating the existence of frequency domain perturbations that are difficult for a human to perceive, yet substantially degrade generalization for DNNs, the authors conclude that DNNs learn surface statistics rather than the abstract concepts that humans rely on. Another example is \citet{Ilyas2019NotBugs}, who argue that the existence of adversarial examples is a natural consequence of the differences between human and machine perception. Finally, \citet{Geirhos2018TextureBias} argue that a key difference between humans and DNNs is that while humans primarily rely on shapes to recognize objects, DNNs are biased towards learning textures.

\section{Experiments}

\paragraph{Dataset.}

We conduct our experiments using the NYU Breast Cancer Screening Dataset~\citep{Wu2019Dataset} comprised of 229,426 screening mammography exams from 141,473 patients. Each exam consists of four images corresponding to the four standard views of screening mammography, where each breast is viewed from two different angles. Each set of images is paired with four labels indicating whether there is a malignant or benign finding in each breast. Henceforth, we refer to these labels as the cancer labels, and use them for training. These precise labels are only available through biopsy, and less than $0.5\%$ of patients in the screening population have cancer. Therefore, we pretrain on exam-level BI-RADS labels which are a radiologist's assessment of the risk of cancer based on screening mammography. The key difference between the cancer labels and the BI-RADS labels are that the latter are not based on biopsy, and thus are more subjective and noisy.

\paragraph{Models.}

In order to draw robust conclusions, we consider two mammogram image classifiers. The first is the Deep Multi-View classifier\footnote{$\nan$ is available online at \url{https://github.com/nyukat/breast_cancer_classifier}.}~\citep{Wu2019Mammography}, which we denote $\nan$. $\nan$ simultaneously takes as input all four images in an exam, and predicts whether there is a malignant or benign finding in each image. This process mimics how radiologists simultaneously utilize all views in an exam when making diagnoses. The second is the Globally-Aware Multiple Instance Classifier\footnote{$\artie$ is available online at \url{https://github.com/nyukat/GMIC}.}~\citep{Shen2019GMIC, Shen2020GMIC}, denoted as $\artie$. $\artie$ takes a single image as its input, and applies a lower capacity network to extract salient patches, which are then processed by a higher capacity network. The information obtained from the two subnetworks are aggregated to predict whether there is a malignant or benign finding in the image. This procedure resembles the way radiologists scan an image to obtain a holistic view, while zooming into particularly suspicious regions. We train both $\nan$ and $\artie$ using the same hyperparameters as the authors.

\begin{figure}[h!]
    \centering
    \includegraphics[width=0.161\textwidth]{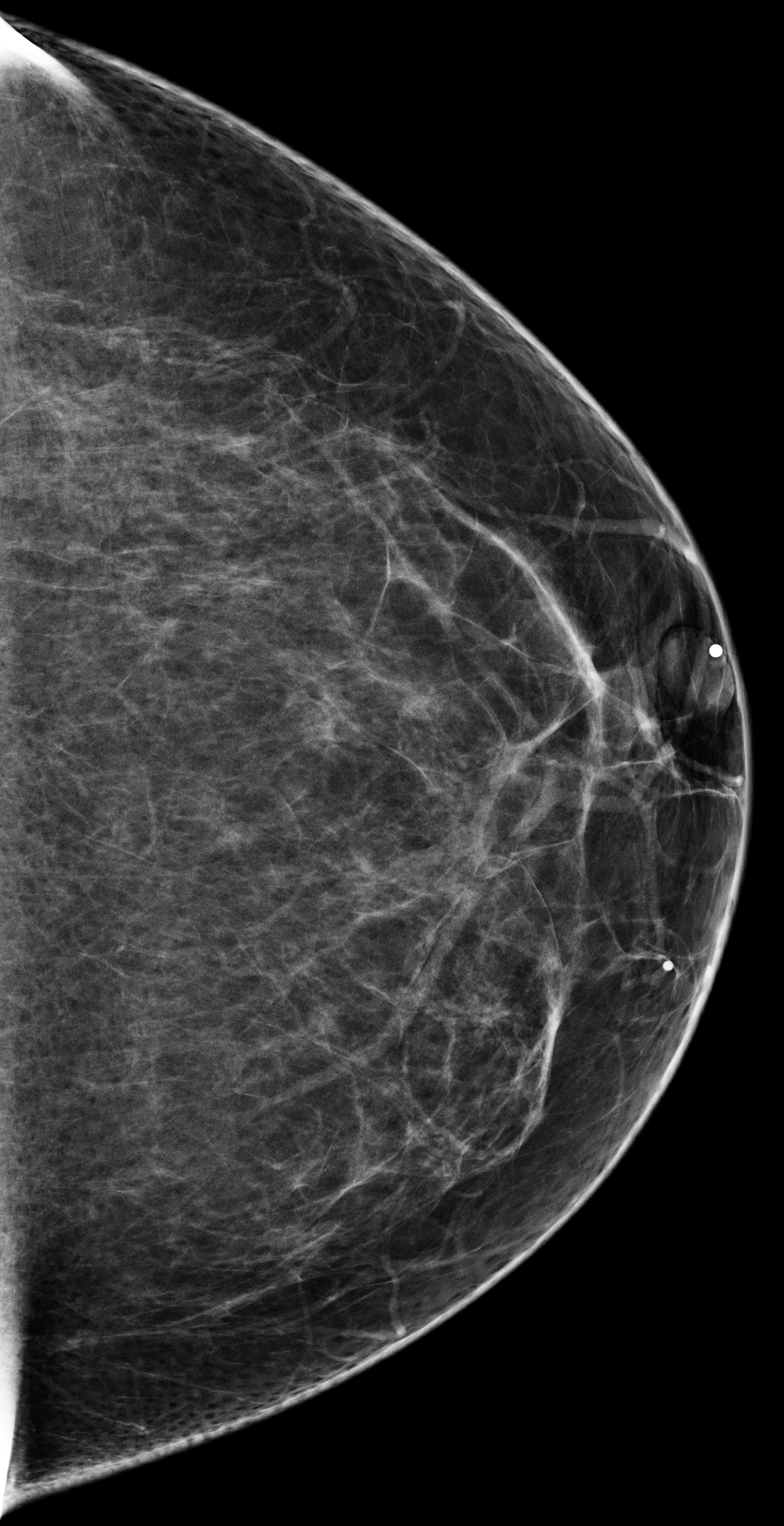}
    \includegraphics[width=0.161\textwidth]{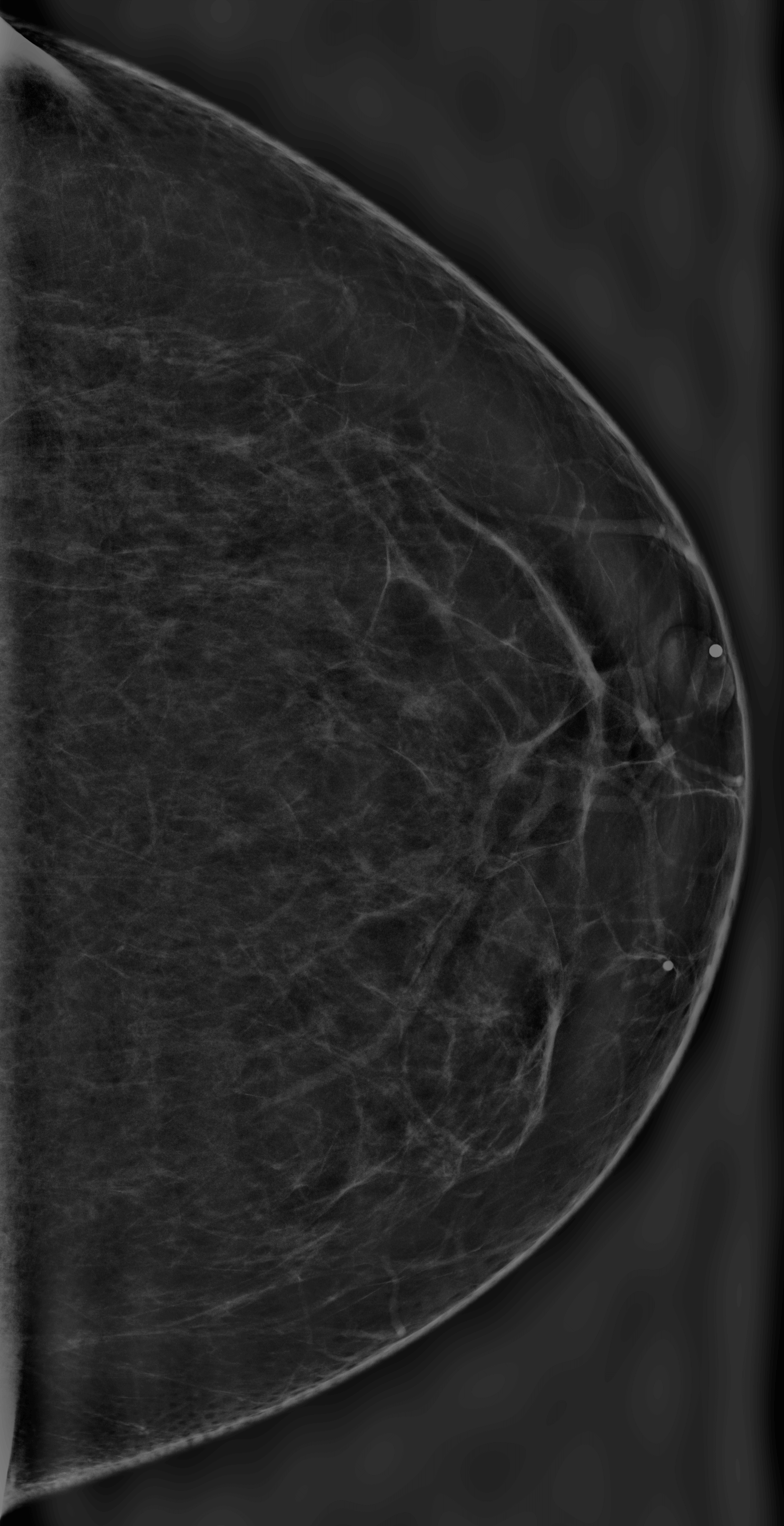}
    \includegraphics[width=0.161\textwidth]{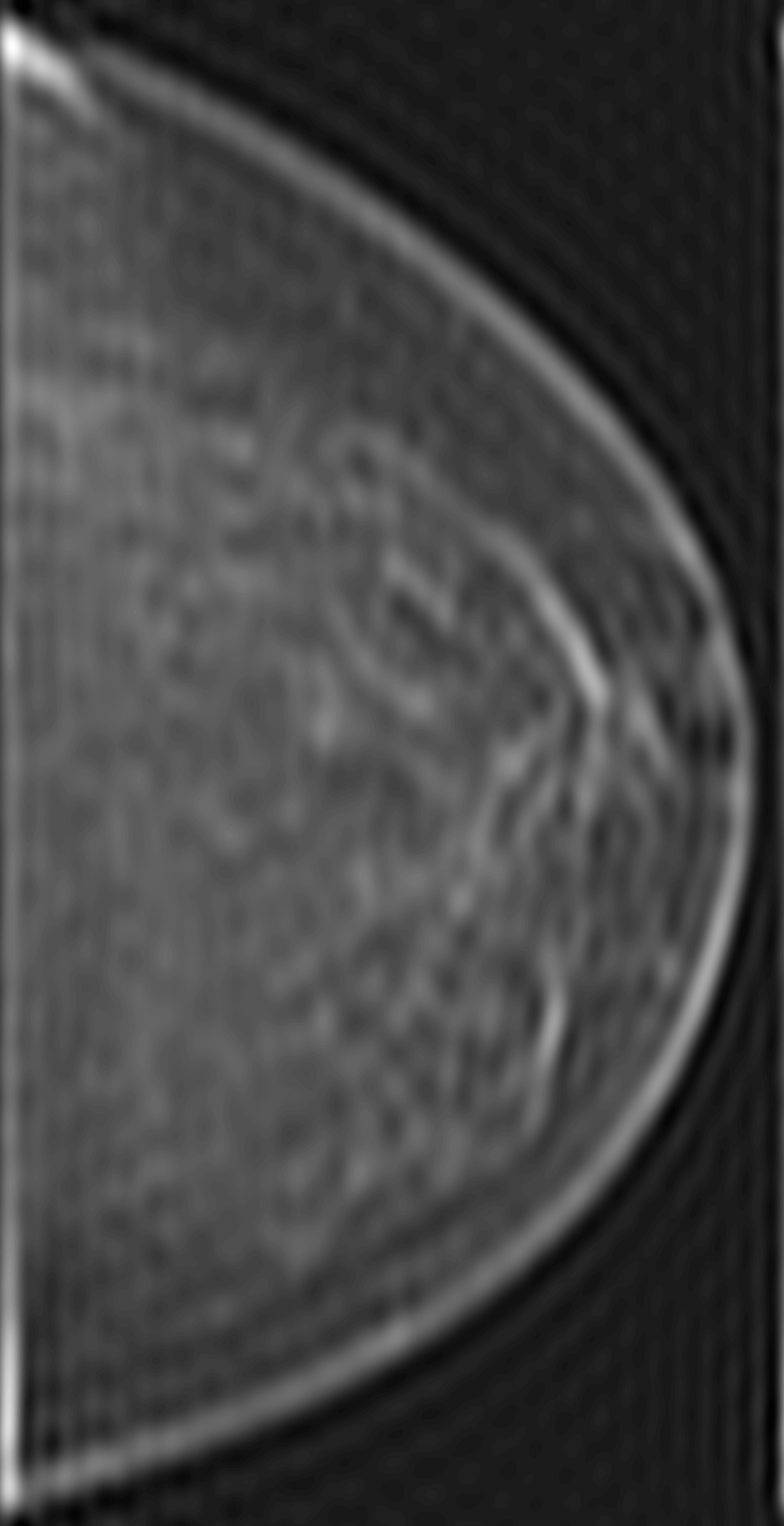}
    \includegraphics[width=0.161\textwidth]{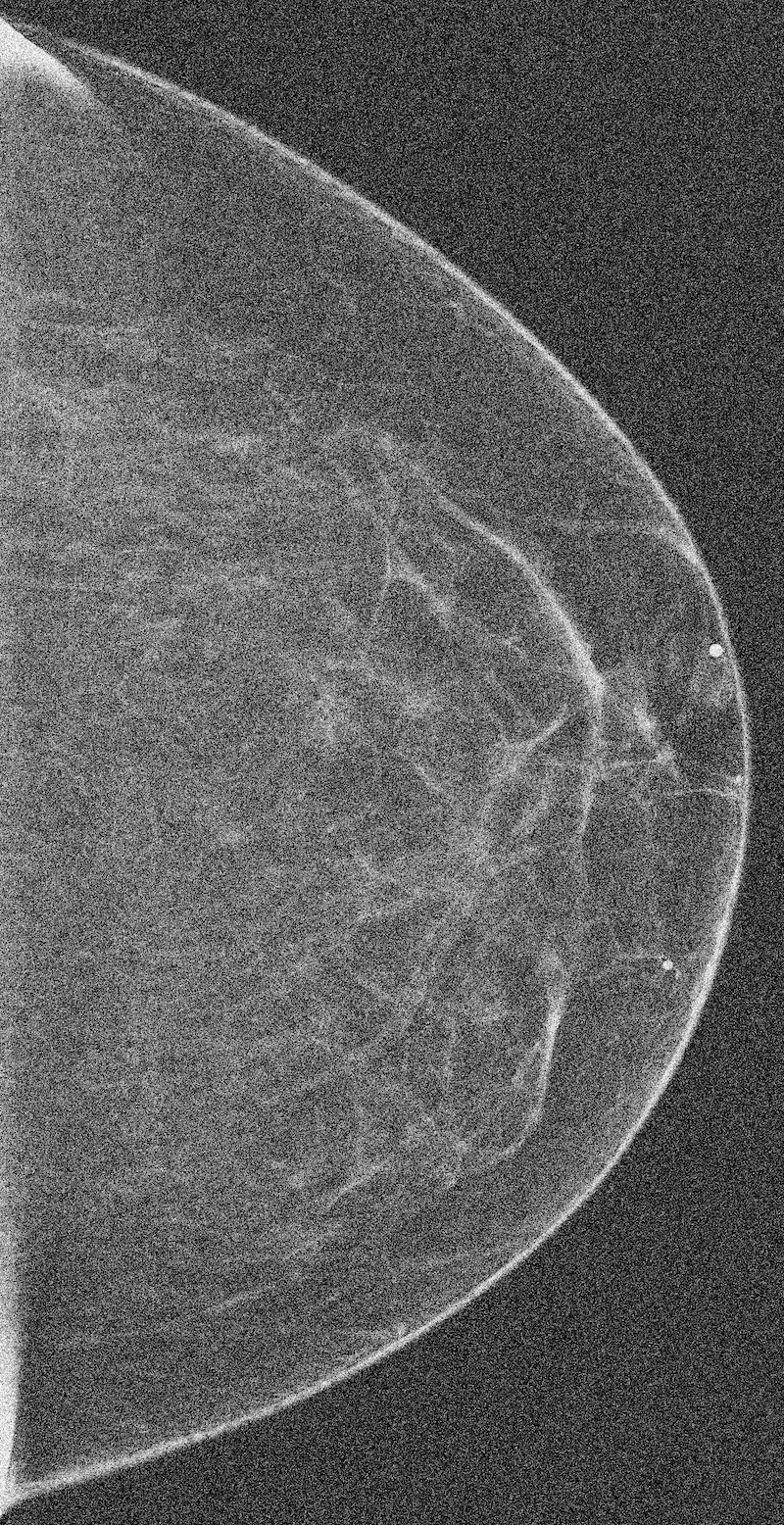}
    \includegraphics[width=0.161\textwidth]{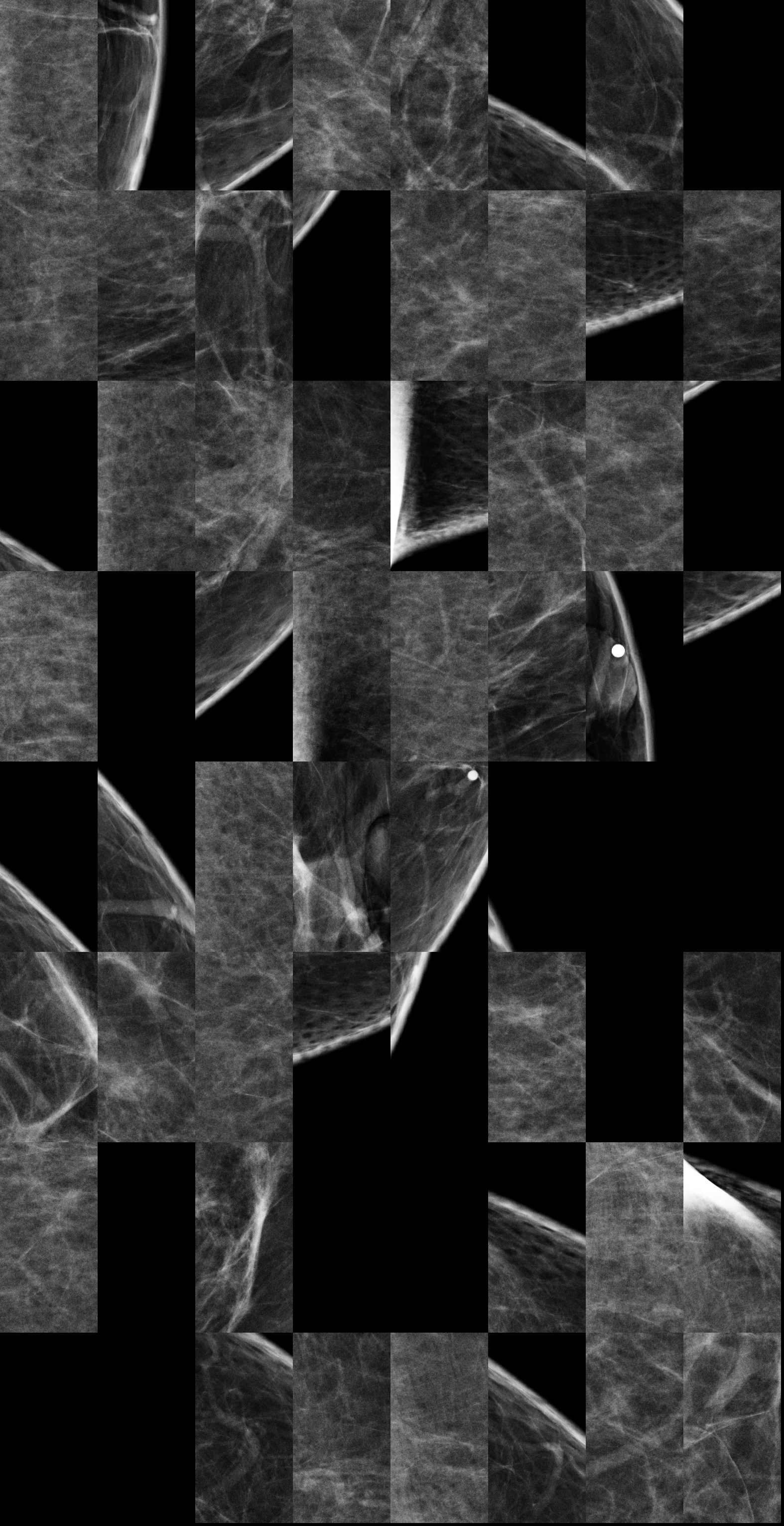}
    \caption{Perturbations applied to an image from the dataset. From left to right: unperturbed image, high-pass filtering, low-pass filtering, Gaussian noise, and patch shuffling.}
    \label{fig:image_perturbations}
\end{figure}

\paragraph{Perturbations.} We evaluate the robustness of each model against four perturbations commonly studied for natural images: high- and low-pass filtering, Gaussian noise, and patch shuffling~\citep{Zhang2019InterpretATCNN}. Figure~\ref{fig:image_perturbations} illustrates these perturbations applied to a sample image from the dataset. In this set of experiments, we apply these perturbations only during inference, and all models are trained using unperturbed data.

The implementation details of the perturbations are as follows. For high-pass filtering, we apply an ideal filter by applying the shifted 2D discrete Fourier transform and attenuating all frequencies lower than $R = 2.0\mathrm{mm}$, measured in cycles per millimeter on the breast. We perform a similar procedure for low-pass filtering, but instead attenuate all frequencies higher than $R = 2.0\mathrm{mm}$. Since the pixels in the images are in the range $[0, 4095]$ prior to normalization, we apply Gaussian noise with mean zero and standard deviation 1600. For patch shuffling, we break the image into 64 square patches and shuffle them. This has the effect of destroying global contextual information that may be useful for breast cancer screening. We also compare each perturbation to a baseline specified as Gaussian noise with standard deviation tuned to match the average $L_2$ norm of the corresponding perturbation.

\subsection{Results}

We begin by evaluating the robustness of the models to the four perturbations, and then take a closer look at the effects of low-pass filtering.

\paragraph{Comparing robustness across perturbations and models.} In Table~\ref{table:1}, we compare the robustness of the models against the four perturbations in terms of the area under the ROC curve (AUC). We report the test set AUC using the weights of the model from the epoch in which it achieved the best AUC on the validation set.

\emph{Our results show that similarly to DNNs trained on natural images, DNNs trained on mammogram images are also vulnerable to high- and low-pass filtering, Gaussian noise, and patch shuffling.}

A natural hypothesis is that sensitivity to individual perturbations stems not from their idiosyncrasies, but from the sensitivity of the model to perturbation in any random direction in the input space. To account for this, we compare each perturbation to a corresponding baseline in the form of Gaussian noise with standard deviation set to match the average norm of the perturbation. We report the performance differences in parentheses in Table~\ref{table:1}. These baselines are distinct from the Gaussian noise results displayed in the table without parentheses, which represent robustness to noise with fixed standard deviation. In terms of this metric, both models perform similarly for high-pass filtering and patch shuffling, but $\artie$ is significantly more robust than $\nan$ against low-pass filtering.

\begin{table}[h!]
\centering
\small
\begin{tabular}{|c|ccccc|}
\hline
Models & \makecell{Unperturbed} & \makecell{High-pass} & \makecell{Low-pass} & \makecell{G. noise} & \makecell{Patch shuffling}\\ [0.5ex] 
\hline
$\nan$ & 0.83 & 0.54 (-0.15) & 0.77 (-0.01) & 0.69 & 0.60 (-0.10)\\
$\artie$ & 0.91 & 0.59 (-0.13)  & 0.87 (+0.10) & 0.73 & 0.70 (-0.08)\\
\hline
\end{tabular}
\caption{AUC of a given model (rows) on the perturbed test sets, as well as on the unperturbed test set (columns). We report in parentheses the difference in AUC compared to a baseline perturbation specified as Gaussian noise with varying standard deviation; please see the text for details. The Gaussian noise results without parenthesis are for a fixed standard deviation of 1600.}
\label{table:1}
\end{table}

\paragraph{A closer look at low-pass filtering.}

\begin{figure}[!h]
    \centering
    \includegraphics[width=0.9\textwidth]{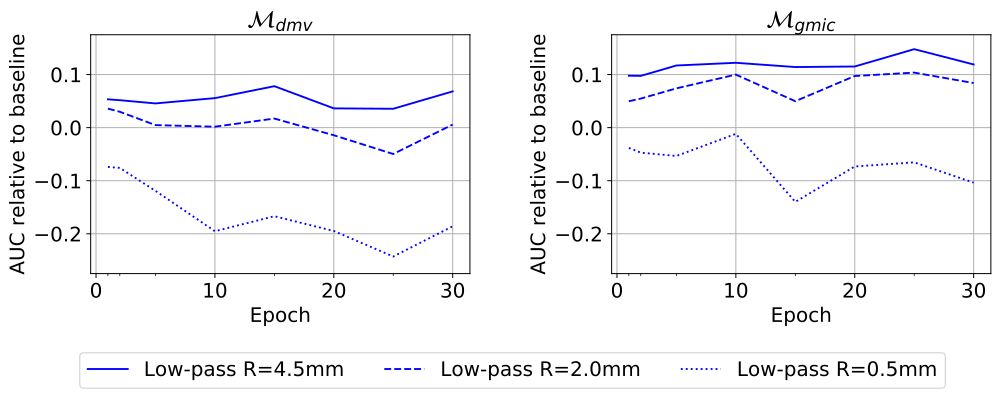}
    \caption{Low-pass filtering applied with various severities at different stages in training for both $\nan$ (left) and $\artie$ (right), where severity is decreasing with respect to $R$.}
    \label{fig:dynamics}
\end{figure}

Several perturbations that are subject to intense study in the context of natural images, such as human-imperceptible adversarial attacks~\citep{Szegedy2014Intriguing}, blurring, and additive noise, are concentrated in high frequencies~\citep{Yin2019Fourier}. These perturbations are particularly interesting because humans are significantly more robust to them compared to DNNs. We therefore concentrate our efforts to understand how mammogram image classifiers are affected by low-pass filtering. Figure \ref{fig:dynamics} illustrates how low-pass filtering affects $\nan$ and $\artie$ at various stages of training. The dynamics remain relatively unchanged from the early stages of training, similar to how test error stabilizes early when classifying low-pass filtered natural images~\citep{Jo2017SurfaceStats}.

\begin{figure}[h!]
\centering
\subfloat[Unperturbed]{
\includegraphics[width=0.18\textwidth]{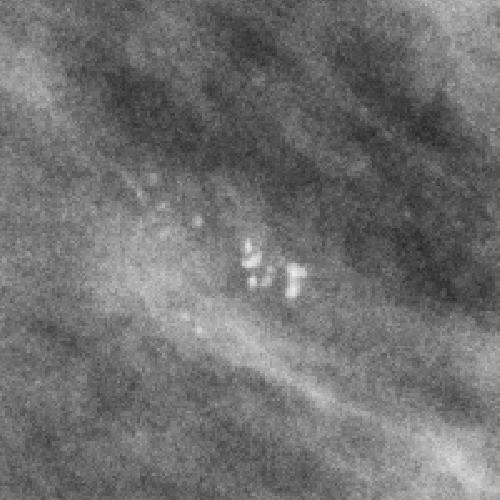}
}
\subfloat[$R = 3.0\mathrm{mm}$]{
\includegraphics[width=0.18\textwidth]{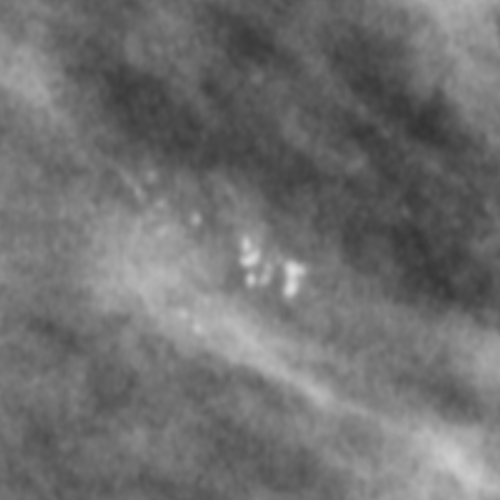}
}
\subfloat[$R = 2.0\mathrm{mm}$]{
\includegraphics[width=0.18\textwidth]{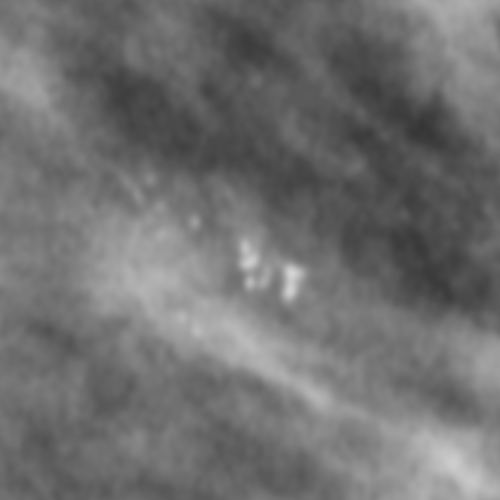}
}
\subfloat[$R = 1.5\mathrm{mm}$]{
\includegraphics[width=0.18\textwidth]{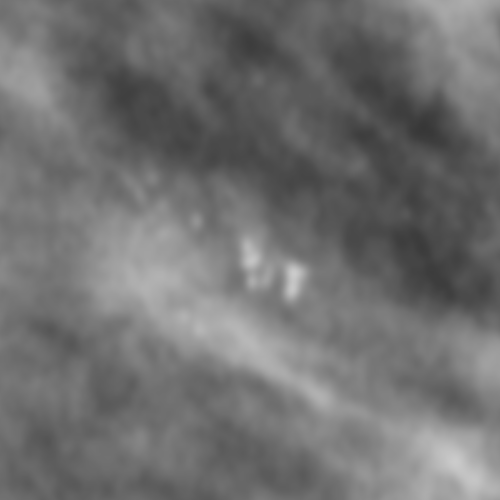}
}
\subfloat[$R = 1.0\mathrm{mm}$]{
\includegraphics[width=0.18\textwidth]{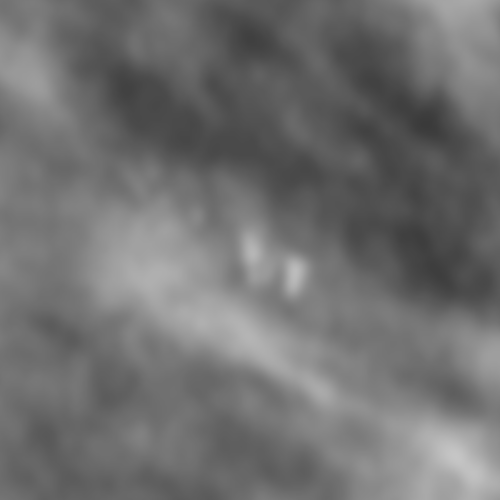}
}
\caption{Mammogram image patch containing microcalcifications, under various severities of low-pass filtering increasing from left to right. Filtering significantly degrades the visibility of this feature.}
\label{fig:microcalc_12}
\end{figure}

\paragraph{Do high frequencies in mammogram images contain features that generalize?} Since low-pass filtering can significantly degrade performance, this leads us to ask whether high frequencies contain valuable information for breast cancer screening. First, we propose an explanation for why high frequencies are important to DNNs. An important potential early symptom of breast cancer are microcalcifications~\citep{Rominger2015}, which are small deposits of calcium in the breast that appear as small white specks on the image. Due to their noise-like appearance, their visibility can be severely degraded by low-pass filtering. We visualize this effect in Figure~\ref{fig:microcalc_12} by applying low-pass filtering with increasing severity to an image patch containing microcalcifications; please see the appendix for more examples. This suggests that high frequencies in mammogram images contain features that are well-established in radiology as being important for breast cancer screening.

\begin{figure}[!h]
    \centering
    \includegraphics[width=0.8\textwidth]{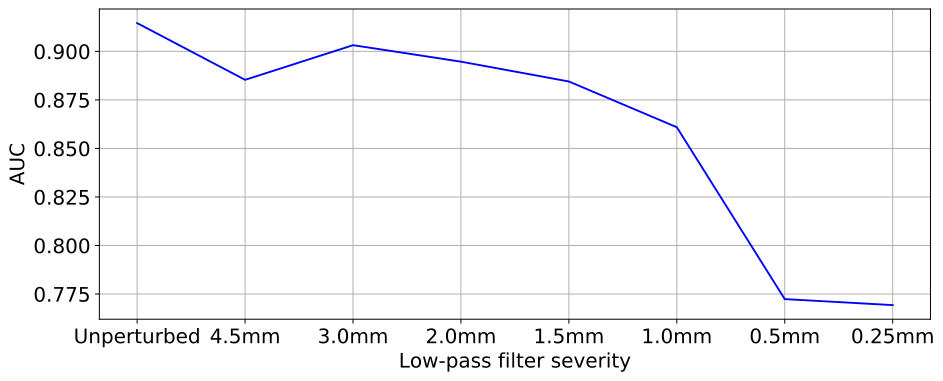}
    \caption{Low-pass filtering applied both during training and inference for $\artie$. The horizontal axis is increasing with respect to the severity of low-pass filtering. Low-pass filtering also degrades performance without the presence of data distribution shift, which suggests that high frequencies carry generalizing features for breast cancer screening.}
    \label{fig:retrain}
\end{figure}

An important consideration is that in the previous experiments, we only applied perturbations during inference. This means that we cannot use the previous results to conclude whether the degradation is attributable to the removal of features that generalize, or to data distribution shift. In order to remove the confounding effect of data distribution shift, we repeat the low-pass filtering experiments for $\artie$, this time applying the perturbations both during training and inference. The results, shown in Figure \ref{fig:retrain}, suggest that high frequencies contain features that generalize in breast cancer screening.

\section{Conclusion}

We empirically evaluated the robustness of two radiologist-level screening mammogram image classifiers to four perturbations commonly studied in the context of natural images. We do not consider these perturbations because of their clinical realism. Instead, our goal is to investigate whether conclusions about robustness drawn from natural images transfer to mammogram images. We found that mammogram image classifiers are sensitive to all four perturbations, which suggests that further work on robustness in screening mammography can build on extensive existing literature. This motivates future research on robustness to more clinically realistic distribution shifts, such as changes in patient demographics. We also took a closer look at low-pass filtering, and identified that it reduces the visibility of clinically meaningful features called microcalcifications. Since these are features that radiologists pay close attention to, it is understandable why corrupting them degrades DNN performance. We additionally argued that invariance to low-pass filtering is undesirable for mammogram image classifiers, while being viewed favorably for natural image classifiers.

A promising direction for future work is to directly compare the sensitivities of radiologists and DNNs to high frequencies in mammogram images. Do DNNs detect human-imperceptible features in the data, or do they detect similar features as radiologists, but are able to utilize them better? Answering these questions will help us apply these promising models in clinical practice.

\newpage

\section*{Acknowledgements}
We thank Florian Knoll, Nan Wu, Yiqiu Shen, and Jungkyu Park for helpful discussions, and Mario Videna and Abdul Khaja for supporting our computing environment. We also gratefully acknowledge the support of Nvidia Corporation with the donation of some of the GPUs used in this research. This work was supported in part by grants from the National Institutes of Health (R21CA225175 and P41EB017183).

\bibliography{main}

\begin{thebibliography}{14}
\providecommand{\natexlab}[1]{#1}
\providecommand{\url}[1]{\texttt{#1}}
\expandafter\ifx\csname urlstyle\endcsname\relax
  \providecommand{\doi}[1]{doi: #1}\else
  \providecommand{\doi}{doi: \begingroup \urlstyle{rm}\Url}\fi

\bibitem[Geirhos et~al.(2019)Geirhos, Rubisch, Michaelis, Bethge, Wichmann, and
  Brendel]{Geirhos2018TextureBias}
Robert Geirhos, Patricia Rubisch, Claudio Michaelis, Matthias Bethge, Felix~A.
  Wichmann, and Wieland Brendel.
\newblock Imagenet-trained cnns are biased towards texture; increasing shape
  bias improves accuracy and robustness.
\newblock In \emph{ICLR}, 2019.

\bibitem[Hendrycks \& Dietterich(2019)Hendrycks and
  Dietterich]{Hendrycks2019CommonCorruptions}
Dan Hendrycks and Thomas~G. Dietterich.
\newblock Benchmarking neural network robustness to common corruptions and
  perturbations.
\newblock In \emph{ICLR}, 2019.

\bibitem[Ilyas et~al.(2019)Ilyas, Santurkar, Tsipras, Engstrom, Tran, and
  Madry]{Ilyas2019NotBugs}
Andrew Ilyas, Shibani Santurkar, Dimitris Tsipras, Logan Engstrom, Brandon
  Tran, and Aleksander Madry.
\newblock Adversarial examples are not bugs, they are features.
\newblock In \emph{NeurIPS}, 2019.

\bibitem[Jo \& Bengio(2017)Jo and Bengio]{Jo2017SurfaceStats}
Jason Jo and Yoshua Bengio.
\newblock Measuring the tendency of cnns to learn surface statistical
  regularities.
\newblock \emph{arXiv}, 1711.11561, 2017.

\bibitem[Krizhevsky et~al.(2012)Krizhevsky, Sutskever, and
  Hinton]{Krizhevsky2012AlexNet}
Alex Krizhevsky, Ilya Sutskever, and Geoffrey~E. Hinton.
\newblock Imagenet classification with deep convolutional neural networks.
\newblock In \emph{NIPS}, 2012.

\bibitem[McKinney et~al.(2020)McKinney, Sieniek, Godbole, Godwin, Antropova,
  Ashrafian, Back, Chesus, Corrado, Darzi, et~al.]{Mckinney2020BreastCancer}
Scott~Mayer McKinney, Marcin Sieniek, Varun Godbole, Jonathan Godwin, Natasha
  Antropova, Hutan Ashrafian, Trevor Back, Mary Chesus, Greg~C Corrado, Ara
  Darzi, et~al.
\newblock International evaluation of an ai system for breast cancer screening.
\newblock \emph{Nature}, 577\penalty0 (7788):\penalty0 89--94, 2020.

\bibitem[Rominger et~al.(2015)Rominger, Steinmetz, Westerman, Ramaswamy, and
  Albert]{Rominger2015}
Marga~B Rominger, Carolin Steinmetz, Ronny Westerman, Annette Ramaswamy, and
  Ute-Susann Albert.
\newblock Microcalcification-associated breast cancer: Presentation, successful
  first excision, long-term recurrence and survival rate.
\newblock \emph{Breast care}, 2015.

\bibitem[Shen et~al.(2019)Shen, Wu, Phang, Park, Kim, Moy, Cho, and
  Geras]{Shen2019GMIC}
Yiqiu Shen, Nan Wu, Jason Phang, Jungkyu Park, S.~Gene Kim, Linda Moy,
  Kyunghyun Cho, and Krzysztof~J. Geras.
\newblock Globally-aware multiple instance classifier for breast cancer
  screening.
\newblock In \emph{MLMI}, 2019.

\bibitem[Shen et~al.(2020)Shen, Wu, Phang, Park, Liu, Tyagi, Heacock, Kim, Moy,
  Cho, and Geras]{Shen2020GMIC}
Yiqiu Shen, Nan Wu, Jason Phang, Jungkyu Park, Kangning Liu, Sudarshini Tyagi,
  Laura Heacock, S.~Gene Kim, Linda Moy, Kyunghyun Cho, and Krzysztof~J. Geras.
\newblock An interpretable classifier for high-resolution breast cancer
  screening images utilizing weakly supervised localization.
\newblock \emph{arXiv}, 2002.07613, 2020.

\bibitem[Szegedy et~al.(2014)Szegedy, Zaremba, Sutskever, Bruna, Erhan,
  Goodfellow, and Fergus]{Szegedy2014Intriguing}
Christian Szegedy, Wojciech Zaremba, Ilya Sutskever, Joan Bruna, Dumitru Erhan,
  Ian~J. Goodfellow, and Rob Fergus.
\newblock Intriguing properties of neural networks.
\newblock In \emph{ICLR}, 2014.

\bibitem[Wu et~al.(2019{\natexlab{a}})Wu, Phang, Park, Shen, Huang, Zorin,
  Jastrz\k{e}bski, F\'{e}vry, Katsnelson, Kim, Stacey~Wolfson, Gaddam, Lin, Ho,
  Weinstein, Reig, Gao, Toth, Pysarenko, Lewin, Lee, Airola, Mema, Chung,
  Hwang, Samreen, Kim, Heacock, Moy, Cho, and Geras]{Wu2019Mammography}
Nan Wu, Jason Phang, Jungkyu Park, Yiqiu Shen, Zhe Huang, Masha Zorin,
  Stanis\l{}aw Jastrz\k{e}bski, Thibault F\'{e}vry, Joe Katsnelson, Eric Kim,
  Ujas~Parikh Stacey~Wolfson, Sushma Gaddam, Leng Leng~Young Lin, Kara Ho,
  Joshua~D. Weinstein, Beatriu Reig, Yiming Gao, Hildegard Toth, Kristine
  Pysarenko, Alana Lewin, Jiyon Lee, Krystal Airola, Eralda Mema, Stephanie
  Chung, Esther Hwang, Naziya Samreen, S.~Gene Kim, Laura Heacock, Linda Moy,
  Kyunghyun Cho, and Krzysztof~J. Geras.
\newblock Deep neural networks improve radiologists' performance in breast
  cancer screening.
\newblock \emph{IEEE Transactions on Medical Imaging}, 2019{\natexlab{a}}.

\bibitem[Wu et~al.(2019{\natexlab{b}})Wu, Phang, Park, Shen, Kim, Heacock, Moy,
  Cho, and Geras]{Wu2019Dataset}
Nan Wu, Jason Phang, Jungkyu Park, Yiqiu Shen, S~Gene Kim, Laura Heacock, Linda
  Moy, Kyunghyun Cho, and Krzysztof~J Geras.
\newblock The nyu breast cancer screening dataset v1.
\newblock Technical report, 2019{\natexlab{b}}.

\bibitem[Yin et~al.(2019)Yin, Lopes, Shlens, Cubuk, and Gilmer]{Yin2019Fourier}
Dong Yin, Raphael~Gontijo Lopes, Jonathon Shlens, Ekin~D. Cubuk, and Justin
  Gilmer.
\newblock A fourier perspective on model robustness in computer vision.
\newblock In \emph{NeurIPS}, 2019.

\bibitem[Zhang \& Zhu(2019)Zhang and Zhu]{Zhang2019InterpretATCNN}
Tianyuan Zhang and Zhanxing Zhu.
\newblock Interpreting adversarially trained convolutional neural networks.
\newblock In \emph{ICML}, 2019.

\end{thebibliography}
\bibliographystyle{iclr2020_conference}

\newpage
\appendix
\label{app:microcalc}
\section{Appendix}

\begin{figure}[h!]
\centering
\subfloat[Unperturbed]{
\includegraphics[width=0.18\textwidth]{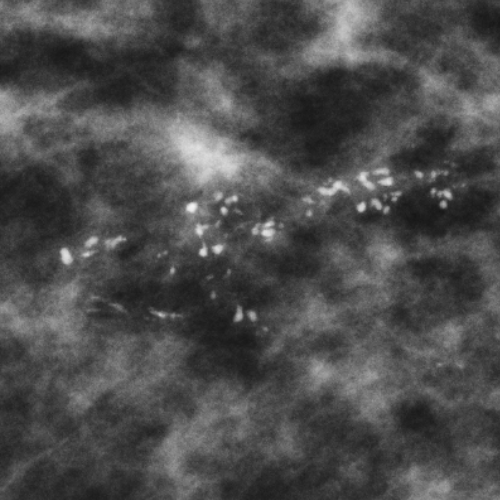}
}
\subfloat[$R = 3.0\mathrm{mm}$]{
\includegraphics[width=0.18\textwidth]{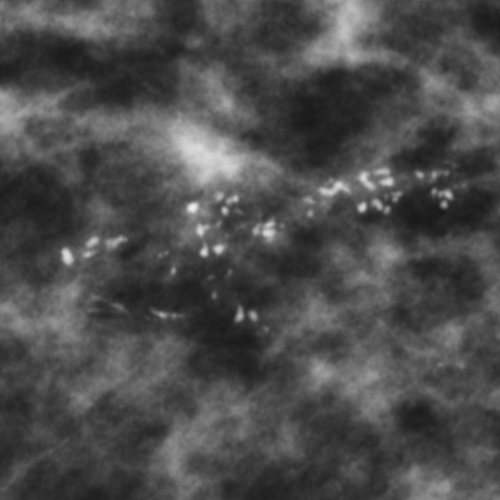}
}
\subfloat[$R = 2.0\mathrm{mm}$]{
\includegraphics[width=0.18\textwidth]{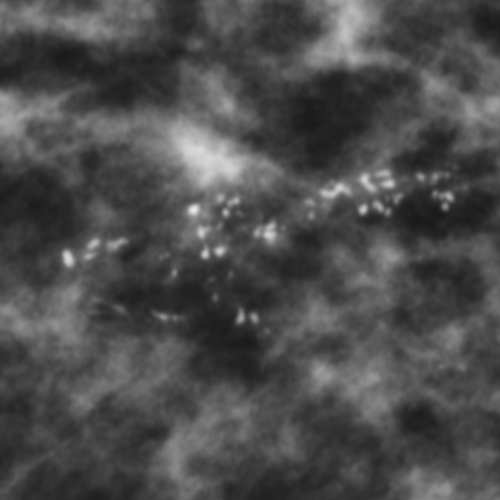}
}
\subfloat[$R = 1.5\mathrm{mm}$]{
\includegraphics[width=0.18\textwidth]{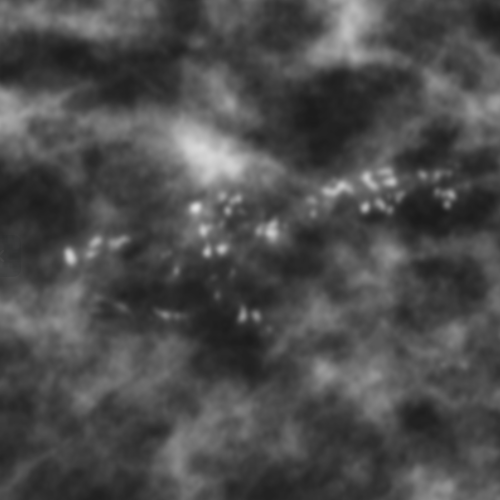}
}
\subfloat[$R = 1.0\mathrm{mm}$]{
\includegraphics[width=0.18\textwidth]{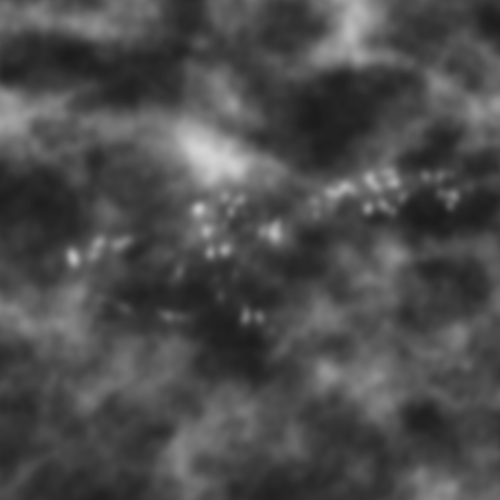}
}
\caption{A mammogram image patch containing strongly visible microcalcifications under various severities of low-pass filtering increasing from left to right. Despite being strongly visible in the unperturbed state, its visibility is significantly diminished under severe filtering.}
\label{fig:microcalc_7}
\end{figure}

\begin{figure}[h!]
\centering
\subfloat[Unperturbed]{
\includegraphics[width=0.18\textwidth]{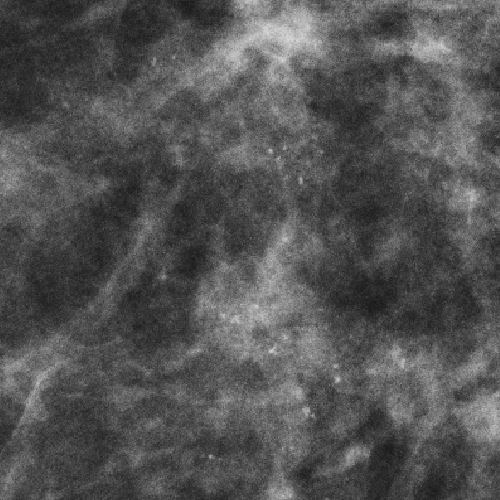}
}
\subfloat[$R = 3.0\mathrm{mm}$]{
\includegraphics[width=0.18\textwidth]{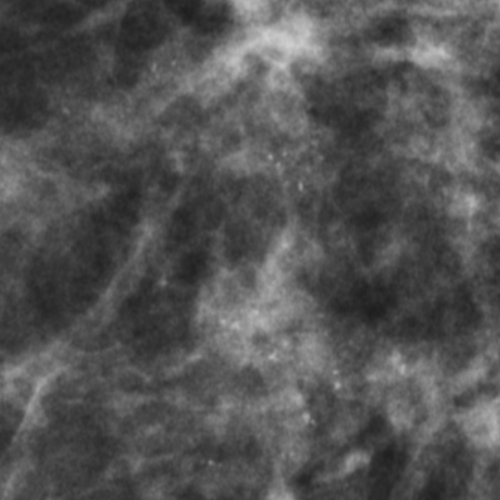}
}
\subfloat[$R = 2.0\mathrm{mm}$]{
\includegraphics[width=0.18\textwidth]{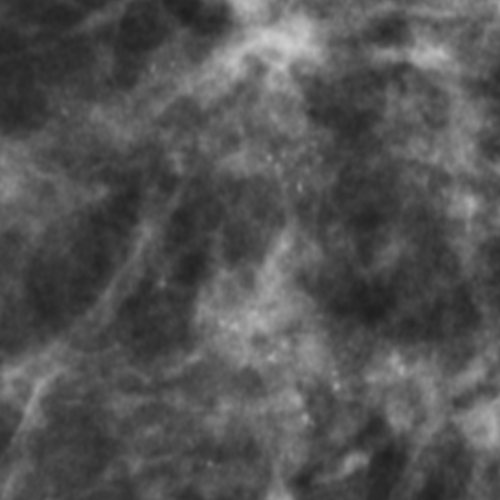}
}
\subfloat[$R = 1.5\mathrm{mm}$]{
\includegraphics[width=0.18\textwidth]{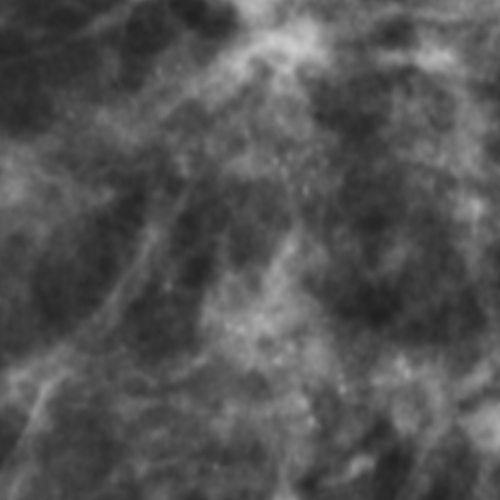}
}
\subfloat[$R = 1.0\mathrm{mm}$]{
\includegraphics[width=0.18\textwidth]{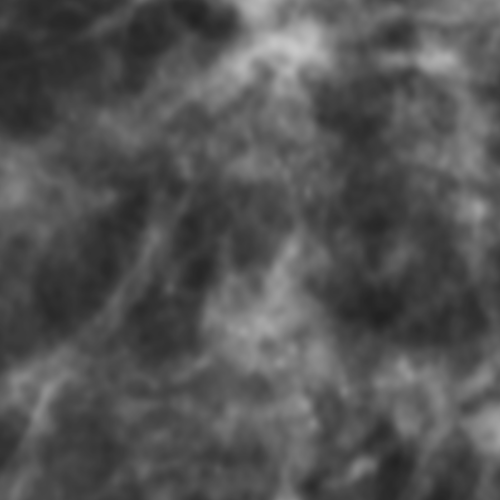}
}
\caption{A mammogram image patch containing weakly visible microcalcifications under various severities of low-pass filtering increasing from left to right. Compared to the previous two examples, the microcalcifications become imperceptible even under mild filtering.}
\label{fig:microcalc_11}
\end{figure}
\end{document}